\documentclass[conference]{IEEEtran}
\usepackage{cite}
\usepackage{amsmath,amssymb,amsfonts}
\usepackage{graphicx}
\usepackage{textcomp}
\usepackage{xcolor}
\usepackage{epsfig}
\usepackage{siunitx}
\usepackage{xspace}
\usepackage{pgf}
\usepackage{caption}
\usepackage{pdfpages}
\usepackage{booktabs}
\usepackage[pagebackref=false,breaklinks=true,letterpaper=true,colorlinks,bookmarks=false]{hyperref}
\usepackage[capitalize]{cleveref}
\usepackage{tabularx,ragged2e}
\def\BibTeX{{\rm B\kern-.05em{\sc i\kern-.025em b}\kern-.08em
    T\kern-.1667em\lower.7ex\hbox{E}\kern-.125emX}}

\graphicspath{{./images/}}

\newcolumntype{L}[1]{>{\raggedright\arraybackslash}m{#1}}
\newcolumntype{R}[1]{>{\raggedleft\arraybackslash}m{#1}}
\newcolumntype{C}[1]{>{\centering\arraybackslash}m{#1}}

\makeatletter

\begin{document}
\title{Facial Behavior Analysis using 4D Curvature Statistics for Presentation Attack Detection}

\author{\IEEEauthorblockN{Martin Th\"{u}mmel, Sven Sickert, and Joachim Denzler}
\IEEEauthorblockA{Computer Vision Group \\
Friedrich Schiller University Jena\\
Jena, Germany \\
\texttt{\{martin.thuemmel,sven.sickert,joachim.denzler\}@uni-jena.de}}
}

\maketitle

%%%%%%%%% ABSTRACT
\begin{abstract}
The human face has a high potential for biometric identification due to its many individual traits.
At the same time, such identification is vulnerable to biometric copies.
These presentation attacks pose a great challenge in unsupervised authentication settings.
As a countermeasure, we propose a method that automatically analyzes the plausibility of facial behavior based on a sequence of 3D face scans.
A compact feature representation measures facial behavior using the temporal curvature change.
Finally, we train our method only on genuine faces in an anomaly detection scenario.
Our method can detect presentation attacks using elastic 3D masks, bent photographs with eye holes, and monitor replay-attacks.
For evaluation, we recorded a challenging database containing such cases using a high-quality 3D sensor.
It features 109 4D face scans including eleven different types of presentation attacks.
We achieve error rates of 11\% and 6\% for APCER and BPCER, respectively.
\end{abstract}

%\begin{IEEEkeywords}
%face, identification, attack, detection, 3D data
%\end{IEEEkeywords}

%%%%%%%%% BODY TEXT
\section{Introduction}
\label{sec:intro}
The digital transformation of organizational processes to improve efficiency in terms of time and resources is progressing fast.
As a result, the demand for automatic and robust unsupervised authentication methods increases.
However, current authentication methods have limitations that make them prone to advanced spoofing attacks or identity thefts.
These \textit{presentation attacks} reduce trust in procedures for automated border control, financial transfer, and mobile payments using self-service eGates, kiosks, and mobile phones.

Biometric methods take advantage of the individuality of human physiological characteristics.
Typical systems are based on finger print, palm print, or iris \cite{zhang2005palm} identification.
Additionally, the field of behavioral biometrics provides methods to analyse voice \cite{cheyer2016device}, walking gait \cite{cattin2002biometric}, and keystroke dynamics \cite{monrose1997authentication}.
By analyzing temporal dynamics these methods can be more robust against presentation attacks than biometric methods that rely on static appearance.
However, individual characteristics can also be replayed or imitated by another person \cite{arik2018neural}.
Therefore, we focus on \textit{presentation attack detection} (PAD) as a mandatory security check for face authentication systems.
We analyze the plausibility of the individual facial trait of a person based on temporal sequences of three-dimensional scans of their face.
In the following, we will refer to these as 4D face scans.

\begin{figure*}[tb]
  \centering
  \includegraphics[width=\textwidth]{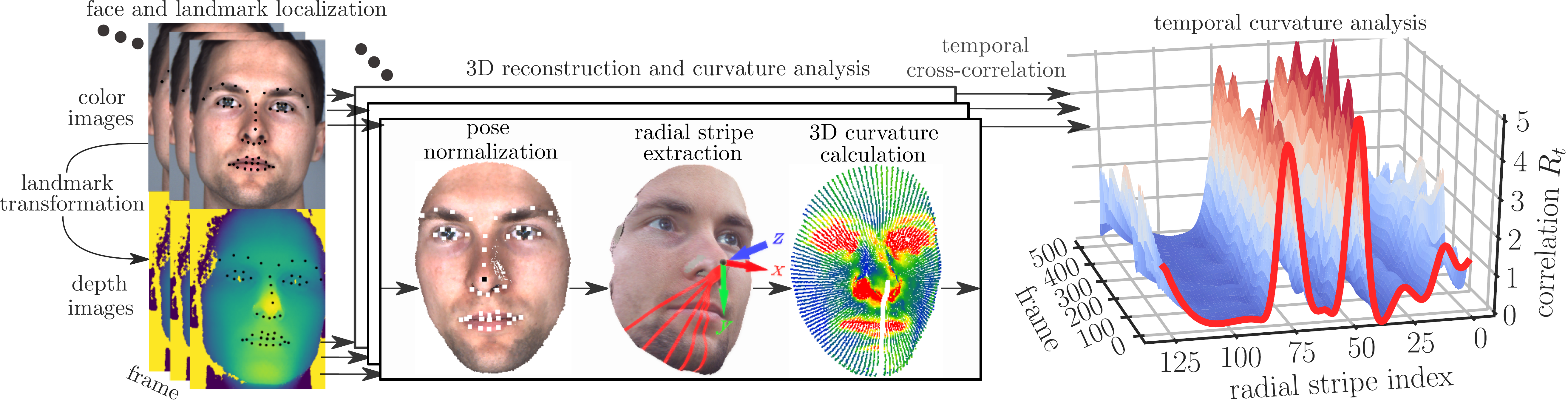}
  \caption{Our suggested representation of 4D face scans reduces the huge amount of biometric characteristics in color and depth image sequences to the graph shown on the right.
    This graph measures the facial expression change at distinct radial stripes over time.
    As such, it is eligible for identifying presentation attacks in an anomaly detection setting using a one-class SVM, for instance.
    This figure is best viewed in color.}
  \label{fig:pipeline}
\end{figure*}

To account for the huge amount of redundant information in high-resolution 4D face scans we analyze the plausibility of curvature changes at sub-sampled radial stripes.
An illustration of the overall pipeline for our curvature analysis is depicted in \cref{fig:pipeline}.
As a result, our novel framework for anomaly detection can detect elastic masks as well as static and planar presentation attacks based on their abnormal facial behavior.
For static and planar presentation attacks, the curvatures are either very small or are constant over time.
The training based only on genuine faces increases generalization capability as properties of individual attack types cannot be memorized. 

\section{Related Work}
\label{sec:relatedWork}
In the following, we first state the developments in the field of presentation attack detection to put our work into perspective.
After that, we look into related work on curvature analysis of radial stripes.

\subsection{Presentation Attack Detection}
\begin{figure}[t]
  \centering
  \includegraphics[width=\linewidth]{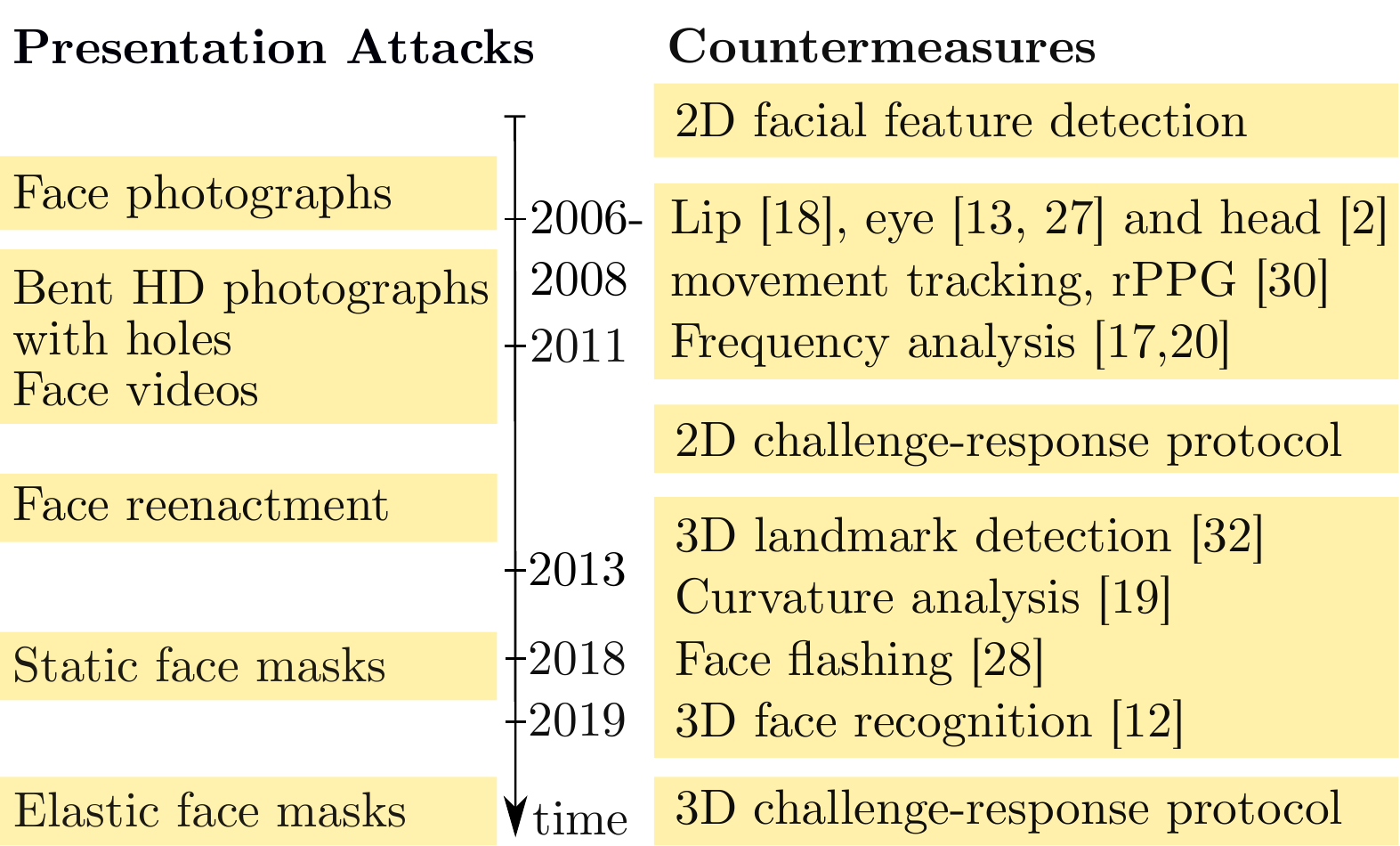}
  \caption{The complexity of presentation attacks increased over time from face photographs to elastic masks. 
    To detect them, a variety of suitable countermeasures were developed.}
  \label{fig:progress}
\end{figure}
In recent years, the quality of presentation attacks included in public 3D face databases continuously improved from photographs to sophisticated silicone masks.
In \Cref{fig:progress}, we give an overview on which kind of biometric copies impostors already used in the past and how they are counteracted.

After accurate 2D face recognition systems were implemented in portable devices, methods for detecting static photographs from differences in Fourier spectra like \cite{li2004live,kim2012face} emerged.
In contrast, these methods are vulnerable to large resolution photographs with high-frequency spectra components.
Additionally, PAD methods based on natural face movements like eye blinking \cite{sun2007blinking} or eye movement \cite{jee2006liveness} can be fooled by additional holes at eyes and mouth.
Thus, 2D challenge-response protocols were introduced, where the user is asked to perform facial expressions \cite{tang2018face}, head movements \cite{bao2009liveness} or read words aloud \cite{kollreider2007real}.
More recently, deep fakes \cite{tewari2018self} became a threat.
They can be used to reenact authenticated faces in real-time on a portable monitor screen as the requested response.

At the same time, cheap 3D sensors become available.
They are integrated into mobile phones using time-of-flight, structured-light or stereo 3D scanning technology.
Due to their high potential for PAD several methods for 3D face scans based on curvature \cite{lagorio2013liveness}, depth variance of 3D landmark locations \cite{wang2013face}, 3D face recognition \cite{tang2018face} have been proposed.
The accuracy of the 3D face recognition method in \cite{gilani2018learning} reached the accuracy of 2D methods and is more robust against planar presentation attacks by design.

Similarly, the authors of \cite{raghavendra2015presentation} are able to detect planar presentation attacks based on facial images using a variety of depth-of-field ranges captured by a light field camera.
However, there are sophisticated static 3D masks or detailed wax figures that can fool said methods.
Hence, we propose to combine a challenge-response protocol with the temporal analysis of 3D face to improve robustness against such sophisticated masks.
In fact, the temporal aspect even allows to detect elastic 3D masks.

As a source for training PAD methods, recent databases contain many wax figures \cite{jia2019spoofing} or realistic silicone masks that are deformed by facial expressions \cite{manjani2017detecting}.
The variety of such masks is limited, such that many 2D PAD methods have already seen the same or similar presentation attacks during training.
Thus, properties of individual attack types might get memorized while methods remain vulnerable to different or unseen attacks.
In contrast, our method requires no samples containing presentation attacks during training.
We propose an anomaly detection scenario, where the PAD method is trained only on recordings of genuine faces.

\subsection{Curvature Analysis of Radial Stripes}
According to \cite{katina2016definitions}, 3D anatomical curves provide a much richer characterization of faces than landmarks, which are just individual points along these curves.
Thus, authors of \cite{vittert2017statistical} localize a complete set of anatomical curves along ridges and valleys of the neutral face shape for statistical analysis.
However, many heuristics would be required to obtain robust anatomical curves in case of facial expressions.
Instead of relying on anatomical curves, the representation proposed in \cite{berretti2008face} approximates the facial shape with a set of geodesics. 
Each so-called radial curve is obtained by intersecting the facial surface with a set of equiangular planes that pass through an anchor point. 
This representation has proven to result in superior performances in 4D facial expression recognition \cite{zhen2017magnifying}, face recognition \cite{drira20133d,berretti2008face}, and even body part analysis in general \cite{bowman2015anatomical,HKUHUB_10722_31166}.

These methods fit 1D polynomial curves in a registration process.
The degree of the polynomials is fixed.
It is assumed that the complexity of the underlying surface is similar for all curves and all expected shapes.
In practice, this results in plausible curvatures values for planar presentation attacks in the same range as for genuine faces due to overfitting.
Therefore, our proposed representation combines the extraction of radial stripes with the calculation of the 3D surface curvature \cite{pauly2002efficient}.
As a larger neighborhood in a 2D surface is taken into account, this approach is more robust with respect to overfitting to noise, outliers, and holes.

\section{Method}
For 4D facial behavior analysis, we have to deal with extremely high spatiotemporal redundancy of 4D face scans.
The differences between both temporally consecutive scans and spatially close point neighbors are very small.
A feature representation should only extract facial expression changes, which constitute the overall facial behavior.
Therefore, we propose a representation of 4D face scans based on equidistant surface curvatures extracted at equiangular radial stripes (\cref{sec:curvature}) and their correlation over time (\cref{sec:temporalRedundancy}).
The latter allows for measuring facial expression changes.
Before modeling, feature extraction and analysis a few pre-processing steps are necessary to ensure a normalized input.

\subsection{Pre-processing Steps}
\label{sec:preprocessing}
Given two synchronized sequences of color and depth images, two pre-processing steps are performed. 
First, $68$ anatomical landmarks are localized in the color images using the method described in \cite{kazemi2014one}, transformed to the disparity map, and 3D reconstructed (see \cref{fig:pipeline}, left).
To also remove the 3D reconstructed background, the point cloud is centered at the nose tip landmark.
Then, all points inside a sphere of radius \mbox{$r = \SI{0.1}{m}$} are extracted.
This removes the large proportion of background in the case of 3D face scans in frontal and cooperative scenarios. 
For pose normalization, a rigid transformation between consecutive 3D landmarks and the localized 3D landmarks in the first face scan is obtained by a Procrustes analysis.
Afterward, these transformations are applied to all temporally consecutive 3D face scans, which in turn proved to be robust against pose changes.

\subsection{Spatiotemporal Curvature Analysis}
\label{sec:curvature}

After 3D reconstruction, normalization and face extraction based on a sphere centered at the nose tip we have a pose normalized face.
A depiction with landmarks (white) and nose tip (black) can be found in the middle block of \cref{fig:pipeline}.
The black dot also serves as an origin of a reference coordinate system given by the three unit vectors $\vec n_j$ in x-, y-, and z-direction, with $j\in\{1,2,3\}$.
Red, green and blue arrows in the middle image point into the positive direction of these axes.
In the next step, the face is intersected with the yz-plane at regular angle steps.
Then, a radial stripe is extracted by taking all points for which the projection onto the x-axis $\vec n_1$ is lower than a certain threshold $\delta$ and the projection onto the y-axis $\vec n_2$ is positive:

\begin{equation}
\left\{ \vec p_i = \begin{pmatrix} x_i \\ y_i \\ z_i \end{pmatrix} : \left| \vec p_i^T \vec n_1 \right| \leq \delta ; \vec p_i^T \vec n_2 > 0 \right\}
\label{eq:stripe}
\end{equation}

among all points $\vec p_i$ of a single 3D face scan.
Afterward, the x- and y-axis are rotated clockwise around the z-axis by $\alpha = \frac{\SI{360}{\degree}}{N}$ and the process is repeated until $N$ radial stripes are extracted.
For instance, the first five out of $N = 24$ red stripes in \cref{fig:pipeline} denote the subset of points that satisfy \cref{eq:stripe} for $\alpha = \frac{\SI{360}{\degree}}{24} = \SI{15}{\degree}$.

To reduce the feature dimension, we calculate the curvature for each sampled point based on its local neighborhood in the original point cloud.
The use of the curvature is encouraged by its invariance under 3D Euclidean transformations.
The curvature can be calculated from the 1D parametric arc-length representation of each curve as

\begin{equation}
\kappa(s) = \frac{y''}{(1 + y'^2)^\frac{3}{2}}.
\label{eq:curvature}
\end{equation}

Resulting 1D curvatures in case of a 3D scan of a flat surface (e.g. face photograph, monitor) tend to be in the range of genuine face scans.
This is due to overfitting of radial curves to noisy point cloud data.
To alleviate this effect, we follow \cite{pauly2002efficient} in approximating the 3D surface curvature using the \textit{surface variation} for each sampled point

\begin{equation}
\tilde \kappa(i) = \left|\frac{\lambda_0}{\lambda_0 + \lambda_1 + \lambda_2}\right|,
\label{eq:surfaceVariation}
\end{equation}

where $\lambda_0 < \lambda_1 < \lambda_2$ and $\lambda_i$ are the eigenvalues of a principal component analysis applied to a neighborhood with a fixed radius of \SI{6}{\milli \meter} in the original point cloud.

In contrast to previous works \cite{berretti2008face,zhen2017magnifying,drira20133d,bowman2015anatomical}, we take the 2D neighborhood into account.
The surface variation is bounded to $\frac{1}{3}$ for isotropically distributed points and avoids the major impact of the unbounded, unstable second derivative in \cref{eq:curvature}.
The 3D curvature calculation step in \cref{fig:pipeline} shows the resulting values obtained by \cref{eq:surfaceVariation} for an exemplary 3D face scan.

To compare temporally consecutive radial stripes, we sub-sample $M$ points along each radial stripe.  
We obtain an equidistant spacing between each point by sampling along their projection to the radial axis of the intersection plane $\vec p_i^T \vec n_i$.
As an important advantage, the degree of equidistant and equiangular sub-sampling can be varied by parameters $M$ and $N$ depending on the desired accuracy and runtime.\\

\subsection{Measuring Facial Expression Change}
\label{sec:temporalRedundancy}
The set of $M$ curvatures for $N$ radial stripes would still be indistinguishable between genuine faces and well-shaped copies like the 3D masks from REAL-f Co.
Therefore, we measure the curvature change between temporally consecutive face scans and perform PAD on the resulting time series representation.
We calculate the point-wise product of the curvature values between temporally consecutive radial stripes.
As the detected position of the nose tip slightly varies, it is necessary to align the radial stripes to each other.
Hence, we compute the maximum cross-correlation as

\begin{equation}
R_t = \max_j \left( \sum_{i=-M}^M\tilde \kappa_t(i) \cdot \tilde \kappa_{t-1}(i + j)\right) 
\label{eq:correlation}
\end{equation}

between radial stripes at time steps $t$ and $t-1$, which refer to consecutive frames in a video.
The maximum cross-correlation measures how similar the temporal curvature values are.
The resulting multivariate time series contains values of $R_t$ for each radial stripe.
It measures the individual change of the facial expression.
For instance, the two inner peaks of the graph for $R_t$ in \cref{fig:pipeline} relate to the eye movements and the two outer peaks to the mouth movements.
Estimating $R_t$ for a high-resolution scan takes \SI{90}{\milli \second} on an Intel Core-i7 CPU.

\begin{figure}[tb]
  \centering
  \input{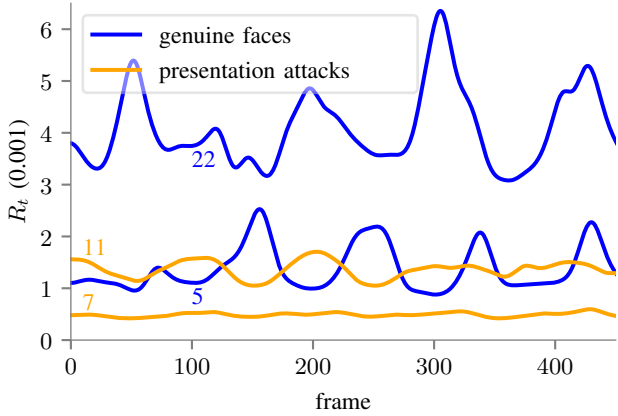}
  \caption{For most recordings of genuine faces like '22', the facial expression change is larger than in case of the 3D mask recording ('7').
    Even when the facial expression change is similar like in '11' (elastic 3D mask) and '5' (minimal facial expression), there is detectable difference in the standard deviation.}
  \label{fig:curvatureChange}
\end{figure}

\begin{figure}[tb]
  \centering
  \input{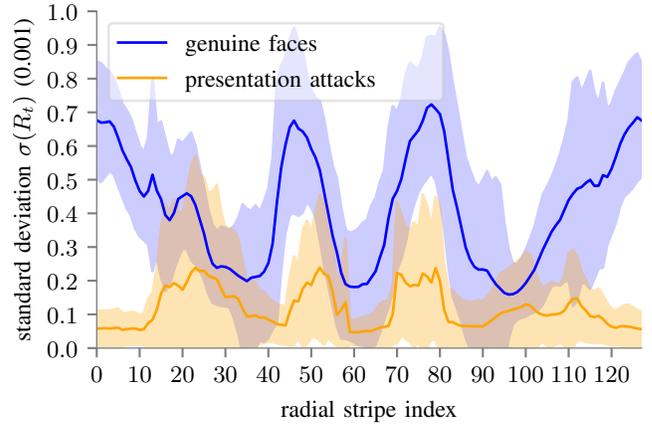}
  \caption{Blue and orange curves show the sample distribution of $\sigma(R_t)$ for each radial stripe ($N = 128$).
    In some cases, standard deviations are small for both presentation attacks and genuine faces.
    However, radial stripes corresponding to mouth and eye regions show substantial differences.}
  \label{fig:curvatureConfidence}
\end{figure}

\cref{fig:curvatureChange} shows some exemplary temporal trends for $R_t$.
It demonstrates that the overall facial expression change of genuine faces is much larger than of presentation attacks using 3D masks.
In contrast, \cref{fig:curvatureConfidence} depicts the distribution of $\sigma(R_t)$ among all radial stripes and recordings in our database (see \cref{subsec:database}).
The standard deviation of the first and last few radial stripes through the mouth differ between genuine faces and presentation attacks by a large margin.
For the eye regions, the standard deviation is also different, but not well-separated.
Current active 3D sensors struggle to accurately measure the eye region, which is either reflecting or absorbing the projected stripe pattern.

In the last step, we need to distinguish changes induced by presentation attacks and changes attributed to genuine faces based on these findings.
Thus, we calculate $\sigma(R_t)$ as a measurement of the overall facial expression intensity for each radial stripe.
In \cref{sec:experiments}, we will report our automatic PAD results based on this feature representation in an anomaly detection setup using a one-class SVM.

\section{Experiments}
\label{sec:experiments}
There are several public databases for PAD that have been used in the past.
However, they do not match our proposed setting for 4D facial behavior analysis.
In some cases, they contain only 2D attacks using bent photographs (CASIA-SURF \cite{zhang2019dataset}) and monitor replay-attacks (CASIA-FASD \cite{zhang2012face}).
Other databases like 3D-MAD \cite{erdogmus2013spoofing}, CS-MAD \cite{bhattacharjee2018spoofing} and WMCA \cite{george2019biometric} containing static, elastic and partial 3D masks only capture the neutral face appearance without any facial expression.
Thus, we created a new database for this evaluation.

\subsection{Database and Protocol}
\label{subsec:database}
\begin{figure*}[tb]
  \centering
  \includegraphics[width=\linewidth]{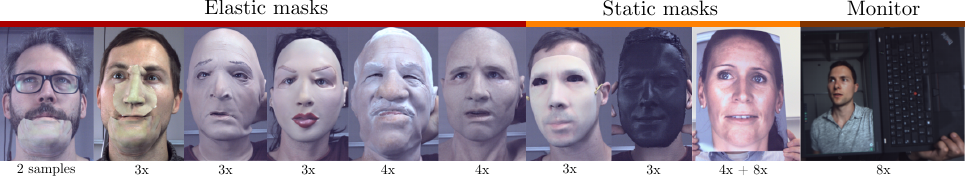} 
  \caption{Our new PAD database contains 4D scans of static presentation attacks, elastic 3D masks and monitors replay-attacks.
    Photographs as static masks were presented without holes (4 samples) and with holes for eyes and mouth (8x).
    Data protection agreements for presenting facial images in compliance with the regulations were signed by all subjects in this paper.}
  \label{fig:database}
\end{figure*}

During the preparations and planning of our recordings, we focused on the cooperative self-service scenario with a frontal pose for face authentication.
Three studio lamps with D65 standard illuminant were used to avoid light flicker and ensure optimal illumination at a constant exposure time of the 3D sensor.
Our application scenarios are the authentication of ID-documents using eGates for cross-border traffic and eKiosks for biometric data enrollment at administrative offices.
Recent mobile phones are also equipped with a 3D sensor.
Hence, the same authentication scenarios can be implemented in mobile devices in the future for mobile payments and contract conclusions.

In addition, we applied a challenge-response protocol.
We asked participants nine questions that reflect common entry regulations, like \textit{Why are you traveling?}, \textit{How long will you stay?} and \textit{Do you carry liquids with you?}
These questions were presented on a video wall behind the 3D sensor.
Participants were instructed to answer orally to induct visemes and facial expressions.

For recording we used a highly accurate structured-light 3D sensor \cite{notni2018irritation}.
Such an active 3D sensor is more exact in homogeneous areas like forehead or cheeks than passive 3D sensors and robust against extraneous light.
Each 3D reconstructed point cloud of a head has $1$ million points. 
On average $250k$ of those points constitute the facial region.
Sequences were recorded at \SI{30}{\Hz} for \SI{36}{\second}, during which questions were asked in \SI{4}{\second} intervals.

In total, our database consist of $109$ 4D face scans from $24$ different subjects (nine women, $15$ men).
Eleven different types of presentation attacks were recorded resulting in $45$ scans.
Furthermore, six participants have beards and six wear glasses.
In the latter case, the participants were recorded with and without glasses.
Beards and glasses pose a challenge for active 3D sensors.
A visual overview with all types of included presentation attacks is given in \cref{fig:database}.
Please note, our database was created in a small study and is not balanced with respect to gender, ethnicity or age.

\subsection{Anomaly Detection Results}
\begin{table}[t]
  \caption{Quantitative evaluation for anomaly detection and standard PAD criteria:
	An SVM is only trained on genuine faces (bona fide) to detect presentation attacks during testing.}
  \label{tab:results}
  \centering
  \begin{tabular}{L{5.8cm} L{2.2cm}}
    \toprule
    \textbf{Performance Measure} & \textbf{Result} \\
    \midrule
    Area under ROC Curve (AUROC) & \SI{95.90}{\percent}\\
    Attack Present. Class. Error Rate (APCER) & $\SI{11.46}{\percent} \pm \SI{3.26}{\percent}$\\
    Bona fide Present. Class. Error Rate (BPCER) & $\SI{6.19}{\percent} \pm \SI{1.23}{\percent}$\\
    \bottomrule
  \end{tabular}
\end{table}

For PAD, we combined the standard deviation of $N = 128$ radial curves in a feature vector for each sample.
Then, we trained an one-class SVM only using genuine faces to pose an anomaly detection task.
Evaluation was carried out using a leave-one-out cross-validation scheme over all subjects.
As parameters of the SVM, we chose a polynomial kernel function and $\nu = 0.05$ as the upper bound of training errors.

\Cref{tab:results} contains the resulting quality metrics on our database described before.
The APCER measures the error rate for presentation attacks wrongly classified as genuine.
BPCER is the analogous measure for \textit{bona fide presentations} (genuine faces).
Both need to be low for a system suitable for practical use.
In our case, we can see a twice as high error rate in the classification performance for presentation attacks when compared to the performance on bona fide presentations.

We observed that the partial chin mask presentations (see \cref{fig:database}, left) particularly pose a challenge for our method.
For the most part, the face is still elastic and able to show normal behavior, which can lead to errors.
Analogously, there are misclassifications in cases of especially strong facial expressions when wearing an elastic mask.
On the other hand, the errors in genuine faces can be traced back to few abnormal recordings.
Corresponding participants did not blink over the whole duration of the recording.
Overall, however, our approach performs very robust in all these different attack scenarios.
We are able to reach an AUC of almost 96\% for this classification task.

\subsection{PAD Discussion}
Given our previous findings, we encourage to specifically localize facial regions, which show anomalous behavior in the next step.
It is possible to highlight the eyes or the partial mask shown in \cref{fig:database} (left) for the chin region.
We also like to point out, that the standard PAD evaluation is only for the classification of genuine faces vs. presentation attacks.
In a more complex task the type of attack could also be classified.

We like to stress, that a limited number of available presentation attack types (e.g. few distinctive masks) increases the potential of bias in a database or algorithm.
However, our anomaly detection approach for PAD does not require training samples of such attacks.
It is not susceptible to overfitting by memorizing properties of mask specifics, which can happen for deep learning-based methods when the training set includes presentation attacks.
The resulting small BPCER shows the generalization capability of our method despite the limited number of subjects in the training set and potential biases from emotional expressiveness, interaction context, or extraversion.

Please note, the steps for pose normalization and curvature analysis preserve the metrical properties of the 3D scans.
Thus, we achieve robustness under head position and pose changes.
Furthermore, the infrared pattern projection of an active 3D sensor results in robustness under illumination changes of visible light.

\subsection{Beyond Presentation Attack Detection}
\begin{figure}[tb]
  \centering
  \raisebox{-0.5\height}{\includegraphics[width=0.85\linewidth]{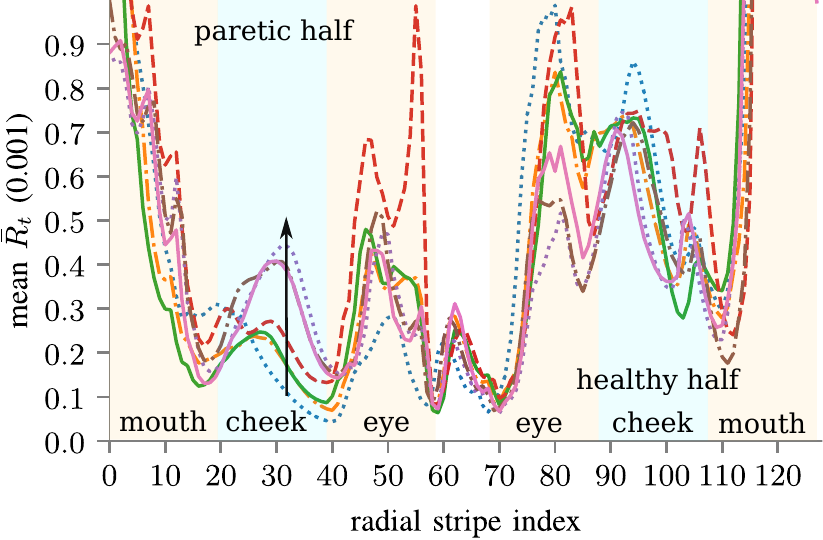}}%
  \hspace*{-0.7cm}
  \raisebox{-0.3\height}{\includegraphics[width=0.2\linewidth]{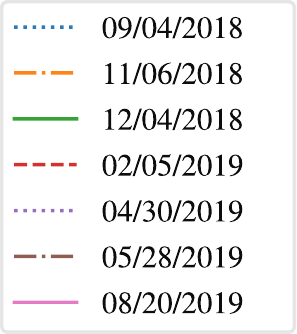}}
  \caption{
    Besides natural fluctuations, $\bar{R}_t$ improved over time for the curves through the left cheek (arrow).
    It indicates, that the treatment method was successful as the nerve reinnervated and allowed for voluntary muscle movements.
    Inner red areas mark radial stripes through the eyes, where recording artifacts can occur.}
  \label{fig:palsyMean}
\end{figure}

Our described scenario for PAD is unique, but is a result of the fast progress in that area (see \cref{fig:progress}).
The approach of time series analysis of 3D point cloud data is not limited to that application scenario.
In a parallel medical study we apply it in paresis treatment analysis.
Patients with facial palsy are recorded at regular time intervals using a similar 3D sensor, but without the challenge-response protocol.
Each individual recording is a temporal sequence of pre-defined facial exercises that serve as training to regain muscle movement.
In this scenario our method serves as a detector and measuring instrument.
It is meant to support physicians in treating such patients as it indicates subtle improvements.
An example is depicted in \cref{fig:palsyMean}, where an improvement of the cheek movement can be observed over time.
The mean of the cross-correlation $\bar{R}_t$ continuously improved.

Beyond this medical task, it is also possible to apply our method for emotion recognition given suitable 4D data and a multi-class classifier.
At the moment typical emotion recognition benchmarks either lack depth information (2D video datasets) or temporal information (single 3D face scans).

\section{Conclusion}
The process of face identification is of ever-increasing importance in a variety of daily scenarios.
At the same time attacks on such systems increase and get more complex.
In this paper, we proposed a PAD method to robustly detect flexible 3D masks, bent photographs, and monitor replay-attacks using 4D face scans and behavioral analysis.
We sub-sampled the 3D surface curvature at equiangular radial stripes and analyzed temporally consecutive stripes to detect anomalies.
Our proposed representation allows for varying the degree of sub-sampling depending on the desired accuracy and run-time.
Furthermore, we created a new dynamic 4D database for the evaluation of our method as available PAD databases do not fit our setting of 4D anomaly detection.
Our results are very promising as the error rates demonstrate.
However, distinguishing elastic 3D masks from genuine faces with minimal facial expressions remains a challenging task.

%\begin{thebibliography}{00}
\bibliographystyle{abbrv}
\bibliography{main}
%\end{thebibliography}

\end{document}